\documentclass[runningheads]{llncs} 

\usepackage{float}
\usepackage{graphicx}
\usepackage{wrapfig}

\usepackage{float}

\usepackage[labelfont=bf]{caption}
\usepackage{amsmath}
\usepackage{algorithm}
\usepackage[noend]{algpseudocode}
\usepackage{verbatim}
\usepackage{color}
\usepackage{amsmath} 
\usepackage{mathtools} 
\usepackage{amsfonts} 
\usepackage{amssymb}
\usepackage[normalem]{ulem}

\usepackage{array}
\usepackage{booktabs}
\usepackage{rotating}
\usepackage{multirow}
\usepackage{multicol}
\usepackage{lscape}
\usepackage{subfig}
\usepackage{comment}

\usepackage{wrapfig}
\usepackage{rotating}
\usepackage{epstopdf}

\newcommand{\squeezeup}{\vspace{-2.5mm}}
\usepackage[export]{adjustbox}

\usepackage{algorithm,algorithmicx,algpseudocode} 

\usepackage{amsmath}

\usepackage{lineno}
\usepackage{soul}
\usepackage{color}
\usepackage{xcolor}
\usepackage{xargs}

\definecolor{brilliantrose}{rgb}{1.0, 0.33, 0.64}
\definecolor{capri}{rgb}{0.0, 0.75, 1.0}
	
\usepackage[colorlinks=true,linkcolor=capri,citecolor=capri]{hyperref}

\usepackage[colorinlistoftodos,prependcaption,textsize=tiny]{todonotes}
\newcommandx{\ISLEM}[2][1=]{\todo[linecolor=brilliantrose,backgroundcolor=brilliantrose!25,bordercolor=brilliantrose,#1]{#2}}

\usepackage{tikz,xcolor,hyperref}

\definecolor{lime}{HTML}{A6CE39}
\DeclareRobustCommand{\orcidicon}{
	\begin{tikzpicture}
	\draw[lime, fill=lime] (0,0) 
	circle [radius=0.16] 
	node[white] {{\fontfamily{qag}\selectfont \tiny ID}};
	\draw[white, fill=white] (-0.0625,0.095) 
	circle [radius=0.007];
	\end{tikzpicture}
	\hspace{-2mm}
}

\foreach \x in {A, ..., Z}{\expandafter\xdef\csname orcid\x\endcsname{\noexpand\href{https://orcid.org/\csname orcidauthor\x\endcsname}
			{\noexpand\orcidicon}}
}
			
\usepackage{color}

\definecolor{darkgreen}{rgb}{0.53, 0.66, 0.42}

\begin{document}

\title{Investigating the Predictive Reproducibility of Federated Graph Neural Networks using Medical Datasets}

\titlerunning{Investigating the Reproducibility of Federated Graph Neural Networks}  

\author{Mehmet Yi\u{g}it Bal{\i}k\index{Bal{\i}k, Mehmet Yi\u{g}it}\inst{1},  Arwa Rekik\orcidB{}\index{Rekik, Arwa} \inst{1,2} \and Islem Rekik\orcidA{} \index{Rekik, Islem}\inst{1}\thanks{ {corresponding author: irekik@itu.edu.tr, \url{http://basira-lab.com}. }}}

\institute{ $^{1}$BASIRA Lab, Faculty of Computer and Informatics Engineering, Istanbul Technical University, Istanbul, Turkey (\url{http://basira-lab.com/}) \\  $^{2}$ Faculty of Medicine of Sousse, Sousse, Tunisia } 

\authorrunning{M.Y. Bal{\i}k et al.}

\maketitle              

\begin{abstract} 
Graph neural networks (GNNs) have achieved extraordinary enhancements in various areas including the fields  medical imaging and  network neuroscience where they displayed a high accuracy in diagnosing challenging neurological disorders such as autism. In the face of medical data scarcity and high-privacy, training such data-hungry models remains challenging. Federated learning brings an efficient solution to this issue by allowing to train models on multiple datasets, collected independently by different hospitals, in fully data-preserving manner. Although both state-of-the-art GNNs and federated learning techniques focus on boosting  classification accuracy, they overlook a critical unsolved problem: \emph{investigating the reproducibility of the most discriminative biomarkers (i.e., features) selected by the GNN models within a federated learning paradigm}. Quantifying the reproducibility of a predictive medical model against perturbations of training and testing data distributions presents one of the biggest hurdles to overcome in developing translational clinical applications. To the best of our knowledge, this presents the first work investigating the reproducibility of \emph{federated GNN models} with application to classifying medical imaging and brain connectivity datasets. We evaluated our framework using various GNN models trained on medical imaging and connectomic datasets. More importantly, we showed that federated learning boosts both the accuracy and reproducibility of GNN models in such medical learning tasks. Our source code is available at \url{https://github.com/basiralab/reproducibleFedGNN}.

\end{abstract}


\section{Introduction}

Over the last years, artificial intelligence (AI) applied to medicine has witnessed exponential growth aiming to ease the diagnostic approach and propel, consequently, the development of personalized treatment strategies. Specifically, advanced deep learning (DL) models such as convolutional neural networks (CNNs) have achieved a remarkable performance across of variety of medical imaging tasks including segmentation, classification, and registration \cite{lee2017deep,shen2017deep}. However, such networks were primarily designed to handle images, thereby failing to generalize to non-euclidean data such as graphs and manifolds  \cite{wolterinkgeometric,bronstein2017geometric}. Recently, graph neural networks (GNNs) were introduced to solve this problem by designing novel graph-based convolutions \cite{bronstein2017geometric,wu2020comprehensive}. A recent review paper  \cite{Bessadok:2021} demonstrated the merits of using GNNs particularly when applied to brain connectomes (i.e., graphs) across  different learning tasks including longitudinal brain graph prediction, brain graph super-resolution and  classification for neurological disorder diagnosis.  Althgouh promising, GNNs remain deep models which are data-hungry.  Faced with the scarcity of medical imaging datasets and their high privacy and sensitivity, they can remain sub-optimal in their performance. In this perspective, federated learning  \cite{mcmahan2017communication} can bring a promising alternative to training GNNs models using decentralized data spread across multiple hospitals while boosting the accuracy of each local GNN model in a fully data-preserving manner. Although increasing the model accuracy through federation seems compelling, there remains a more important goal to achieve which is maximizing the \emph{reproducibility} of a locally trained model. A model is defined as highly reproducible when its top discriminative features (e.g., biomarkers) remain unchanged against perturbations of training and testing data distributions as well as across other models \cite{nebli2022quantifying,georges2020identifying,georges2018data}. Quantifying the reproducibility of a predictive medical model presents one of the biggest hurdles to overcome in developing translational clinical applications. In fact, this allows identifying  the most \emph{reproducible biomarkers} that can be used in treating patients with a particular disorder. To the best of our knowledge, reproducibility in federated learning remains an untackled problem.

\cite{nebli2022quantifying} proposed the first framework investigating the reproducibility of GNN models. Specifically, the designed RG-Select framework used 5 different state-of-the-art GNN models to identify the most reproducible GNN model for a given connectomic dataset of interest. Although RG-Select solves both GNN reproducibility and non-euclidean data learning problems, it does not address the problem of model reproducibility when learning on decentralized datasets distributed across different hospitals. Undeniably, medical datasets carry information about patients and their medical conditions. Hence, the patient may be identified using such data. Patients have the right to control their personal information and keep it for themselves \cite{forcier2019integrating}. Such data must be held private between the patient and their health care workers. For such reasons, federated learning presents a great opportunity to learn without clinical data sharing and while boosting the model accuracy as well as its reproducibility.

We draw inspiration from the seminal work on decentralized learning where \cite{mcmahan2017communication} proposed a federated averaging algorithm based on training many local models on their local datasets then aggregating the learned models at the server level. Next, the global server broadcasts their learned weights to each local modal for local updates. Several researchers were inspired by federated learning and adapted it to graphs \cite{chen2021fedgl,he2021fedgraphnn}. Even though these proposed frameworks managed to boost the local accuracy of local models while handling decentralized data, they overlook the reproducibility of the most discriminative features (i.e., biomarkers). Will federated learning also boost the reproducibility of locally trained GNN models? Here we set out to address this prime question by quantifying the reproducibility of federated local models.


In order to ensure high accuracy, handle decentralized datasets and identify the most reproducible discriminative features, we federate GNN models and quantify their reproducibility by perturbing training and testing medical data distributions through random data splits. Our framework generalizes the seminal work of RG-Select \cite{nebli2022quantifying} to federated models. Specifically, given a pool of  GNN architectures to federate, we aim to identify the most reproducible GNN model across local hospitals and its corresponding biomarkers by quantifying the reproducibility of the global model. The key contributions of our framework are to: \textbf{(1)} Federate the learning of predictive GNN models with application to medical imaging and connectomic datasets.
\textbf{(2)} Investigate and quantify the reproducibility of federated GNN models, and \textbf{(3)}  identify the most \emph{reproducible} biomarkers for neurological disorder diagnosis.

\squeezeup
\squeezeup

\section{Proposed Method}

\begin{figure}[ht!]
\centering
\includegraphics[width=12.5cm]{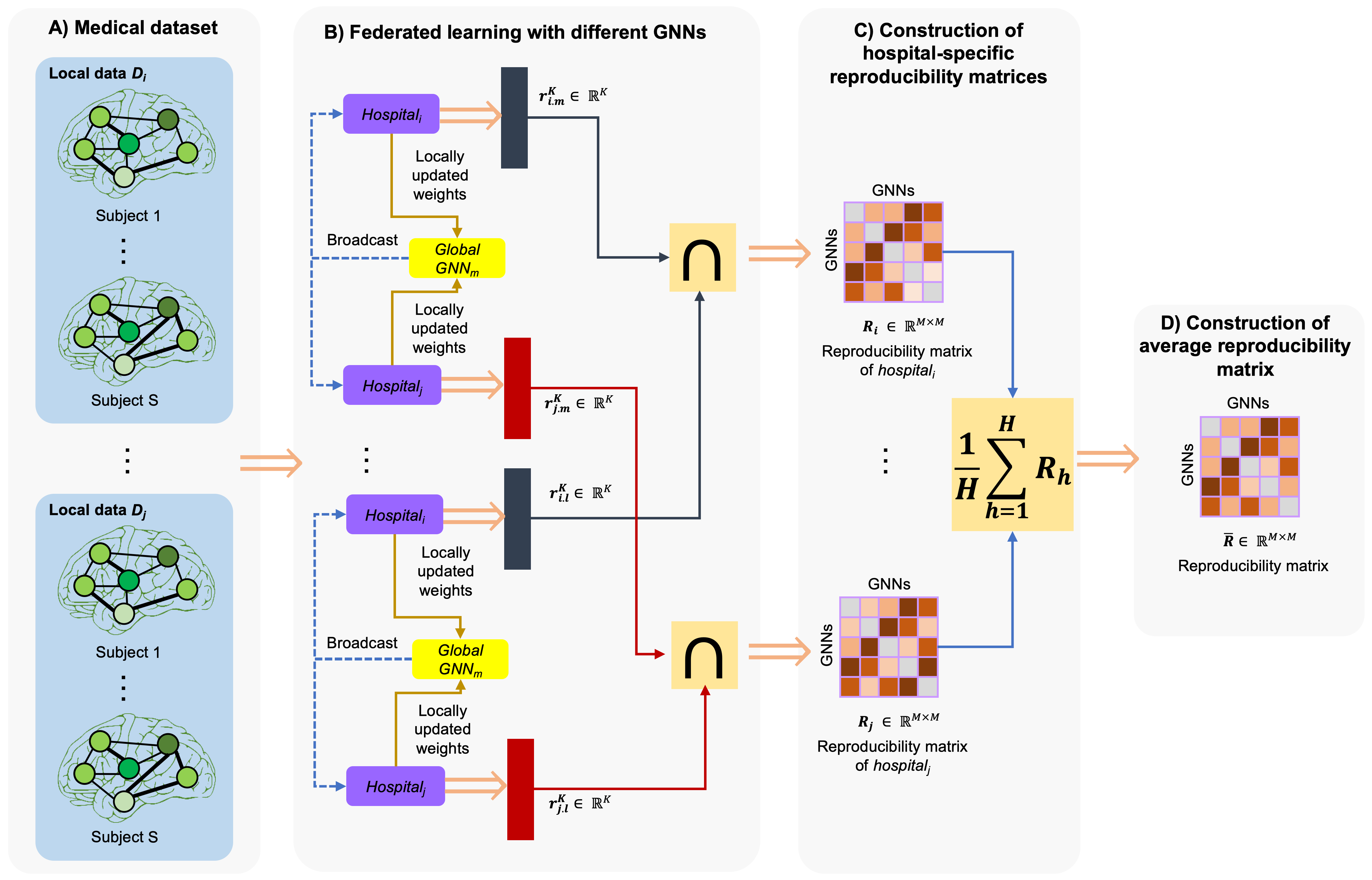}
\caption{\emph{Overview of the proposed framework for quantifying the reproducibility of federated GNN models across decentralized datasets.} \textbf{(A) Medical datasets.} We split our dataset into $H$ local datasets. \textbf{(B) Federated learning with different GNNs.} We use $M$ GNN models to identify the most reproducible GNN model during the federation learning. For each local hospital model $GNN_m$ where $m \in \{1, \dots, M\}$, we extract its top $K$ discriminative features and calculate their overlap ratio with discriminative feature sets selected by other GNN models. \textbf{(C) Construction of hospital-specific reproducibility matrix.} Using the intersections calculated in the previous step, we construct the hospital-specific reproducibility matrix where each element $(i,j)$ denotes the overlap in the top $K$ features identified by the locally trained $GNN_i$ and $GNN_j$. \textbf{(D) Construction of average global reproducibility matrix across federated models.} Using the produced hospital-specific reproducibility matrices, we calculate the average global reproducibility matrix, thereby identifying the most reproducible features across models and hospitals.}
\label{fig:main} 
\end{figure}

In this section, we detail our federated reproducibility quantification framework as illustrated in \textbf{Fig}~\ref{fig:main}. First, we divide the whole data into $H$ different subsets. Each subset represents the local data of a particular hospital. Second, we train different GNN models using federated learning trained on each local dataset. Following the training, we extract the top $K$ discriminative biomarkers (features) identified by each locally trained GNN model. Next, for each hospital, we produce a \emph{hospital-specific GNN-to-GNN reproducibility matrix} where each element denotes the overlap ratio between the extracted top $K$ biomarker sets by pairs of locally trained GNN models. We then construct the \emph{global reproducibility matrix} by averaging all hospital-specific reproducibility matrices. Finally, we identify the most reproducible GNN model across hospitals in the federation process by identifying the central node with the highest overlap with other nodes in the global average reproducibility matrix. The selected model is then used to identify the most reproducible features.

\textbf{Problem statement.}
Given $H$ hospitals with the local datasets $\mathcal{D}_h=(\mathcal{G}_h,\mathcal{Y}_h)$ that belongs to the $h^{th}$ hospital, where $h \in \{1, 2, \dots ,H \}$, let $\mathcal{D}_h$ denote a local dataset including subjects with their diagnostic states/labels (e.g., normal control and disordered). Let $S$ denote the number of subjects in $\mathcal{D}_h$. $\mathcal{G}_h = \{\mathbf{G}_{h,1}, \mathbf{G}_{h,2}, \dots ,\mathbf{G}_{h,S}\} $ denotes the set of medical data graphs and their labels are denoted by $\mathcal{Y}_h=\{y_{h,1}, y_{h,2}, \dots ,y_{h,N}\}$. Each graph $\mathbf{G}_{h,n}$ is represented by an adjacency matrix $\mathbf{X}_{h,n} \in \mathbb{R}^{N \times N}$ and a label $y_{h,n} \in \{0, 1\}$ where $N$ denotes the number of brain regions of interest (ROIs) for connectivity datasets or pixels for medical imaging datasets. Note that $N$ also represents the number of nodes in the corresponding graph.

Given a pool of $M$ GNNs $\{{GNN}_1, {GNN}_2, \dots {GNN}_M\}$, we are interested in training a GNN model ${GNN}_{h,m}: \mathcal{G}_h \rightarrow \mathcal{Y}_h$ on the local dataset of hospital $h$. Our aim is to identify the most reproducible biomarkers or features that discriminate between the two classes. Hence, we extract the top $K$ features $r_{h,m}^{K} \in \mathbb{R}^{K}$ learned by the $m^{th}$ local GNN model in the $h^{th}$ hospital, where $m \in \{1, 2, \dots ,M \}$. We calculate the intersection of the extracted local top $K$ features $r_{h,m}^{K} \cap r_{h,l}^{K}$, where $m$ and $l$ are the indexes of GNN models in the GNN pool and $h$ is the index of a hospital. In order to calculate the reproducibility matrices, we extract the weights $\mathbf{w}_{h,m} \in  \mathbb{R}^{N}$ learned by the $h^{th}$ hospital using the $m^{th}$ GNN architecture.

\textbf{\emph{Definition 1.}} \emph{Let $GNN_i$ and $GNN_j$ be two GNN models and let $\mathbf{w}_i \in \mathbb{R}^{n}$ and $\mathbf{w}_j \in \mathbb{R}^{n}$ be their weights, respectively. The top $K$ biomarkers extracted using the weights $\mathbf{w}_i, \mathbf{w}_j$ are denoted by $r_i^{K}$ and $r_j^{K}$, respectively.
Reproducibility among models $GNN_i$ and $GNN_j$ is denoted by $\mathbf{R}^{K}_{i,j}$ which can be calculated as: $\mathbf{R}^{K}_{i,j} = \frac{|r_i^{K} \cap r_j^{K}|}{K}$}.

\textbf{GNN training mode.} Each local data is divided into 3 folds where 2 folds are used for training and the left-out fold is used for validation. We train each local GNN on its local dataset over $E$ epochs and using $B$  batches. Both global and local models communicate for $C$ rounds. In each round, the global model sends a deep copy of the current GNN model to all local hospitals. Each hospital does training using its local data. When the training ends, hospitals send locally updated weights to the central server. The server applies \textbf{Algorithm}~\ref{FedAvgAlgorithm} on the weights that came from the local models and loads the averaged weights to the global model.

\textbf{Biomarker selection.} We extract the learned weights by each GNN model in order to select the top $K$ discriminative biomarkers. The extracted weights belong to the last embedding layer of the GNN model \cite{nebli2022quantifying}. Next, we rank the biomarkers according to the absolute value of their corresponding weights and select the top $K$ with the highest weights. We use these biomarkers to construct GNN-to-GNN hospital-specific reproducibility matrices.

\begin{algorithm}[ht]
\caption{FederatedAveraging. $H$ hospitals indexed by $h$; $C$ is the number of communication rounds; $G$ is the global model}\label{FedAvgAlgorithm} 
\begin{algorithmic}[1]
\State $\textbf{LocalUpdate($G$)}: $ // Runs on hospital $h$
\For {each epoch $i$ in \{1, \dots , $E$\}}
\For {batch $b$ in $B$}
\State $w \gets w - \eta \nabla l(w;b)$
\EndFor
\EndFor
\State $\textbf{return}$ $w$
\State $\textbf{Server Executes:}$
\State initialize global model $G$
\For {each round $t$ in \{1, \dots , $C$\}} 
\For {each hospital $h$ in \{1, \dots , $H$\}} 
\State $w_{t + 1}^{h} \gets$ LocalUpdate(deepCopy($G$)) // Copy of global model sent to local update
\EndFor
\State $w_{t+1} \gets \sum_{h=1} ^{H} \frac{w_{t+1}^{h}}{H}$
\EndFor
\end{algorithmic}
\end{algorithm}
\begin{algorithm}[ht]
\caption{AvgRepMatrixConstruction. $W$ weights of all GNNs; $K$ is the threshold value}\label{ReproducibilityConstrucAlgorithm} 
\begin{algorithmic}[1]
\State $\textbf{RepMatrixConstruction}$($W, K$):
\For{$w_i$ and $w_j$ in $|W|$} // absolute value of weights is used
\State $ r_i^{K} \gets$ Top $K$ features from $w_i $ 
\State $ r_j^{K} \gets$ Top $K$ features from $w_j $
\State $\mathbf{R}^{K}_{ij} \gets \frac{|r_i^{K} \cap r_j^{K}|}{K}$
\EndFor
\State $\textbf{return}$ $\mathbf{R}^{K}$
\State $\textbf{Execute:}$
\State initialize $\bar{R} \in R^{M \times M}$ with zeros
\For {each hospital weights $W_{h}$ where $h$ in \{1, \dots , $H$\}}
\State $\bar{R} \gets \bar{R} + $ RepMatrixConstruction($W_{h}, K$) 
\EndFor
\State $\bar{R} \gets  \frac{\bar{R}}{H}$
\end{algorithmic}
\end{algorithm}

\textbf{GNN-to-GNN reproducibility matrix.}
Using the top $K$ biomarkers, the overlap of each pair of GNN models is calculated thereby producing their GNN-to-GNN reproducibility score. This step is executed for each hospital individually to produce a hospital-specific reproducibility matrix. Repeating this operation for all $H$ hospitals, the average of $H$ hospital-specific matrices is then calculated, and the average global reproducibility matrix is constructed using \textbf{Algorithm}~\ref{ReproducibilityConstrucAlgorithm}.

\textbf{The most reproducible GNN and biomarker selection.} In order to select the most reproducible GNN model, we use the average reproducibility matrix of the $H$ hospital-specific reproducibility matrices. We consider this matrix as a graph where the GNN models are its nodes. We use the highest node strength to identify the most reproducible global federated model (\textbf{Definition 2}). In fact, such a hub GNN node implies a maximal overlap with other GNN models, thereby evidencing its reproducible power. Next, we find the most reproducible $K$ biomarkers with the highest weights learned by the most reproducible GNN model.

\textbf{\emph{Definition 2.}} Given $M$ GNN models to federate, let $\mathbf{R} \in \mathbb{R}^{M \times M}$ denote the constructed reproducibility matrix where each element encodes the intersection rate of the top $K$ biomarkers identify by pairs of global GNN models. Let $\mathbf{r}_i$ be the $i^{th}$ row of $\mathbf{R}$ where $i \in \{1, 2, \dots ,M \}$. The $\mathbf{r}_i$ includes the top $K$ biomarkers intersection ratios of $GNN_i$ with all GNN models including itself. Let $s_i$ denote the strength (i.e., score) of $GNN_i$ defined as: $s_i = (\sum_{m=1} ^{M} \mathbf{r}_{i,m}) - 1$ (minus one is for excluding the relation with itself).

\begin{figure}[htp!]
\captionsetup{belowskip=0pt}
\centerline{\includegraphics[width=17cm]{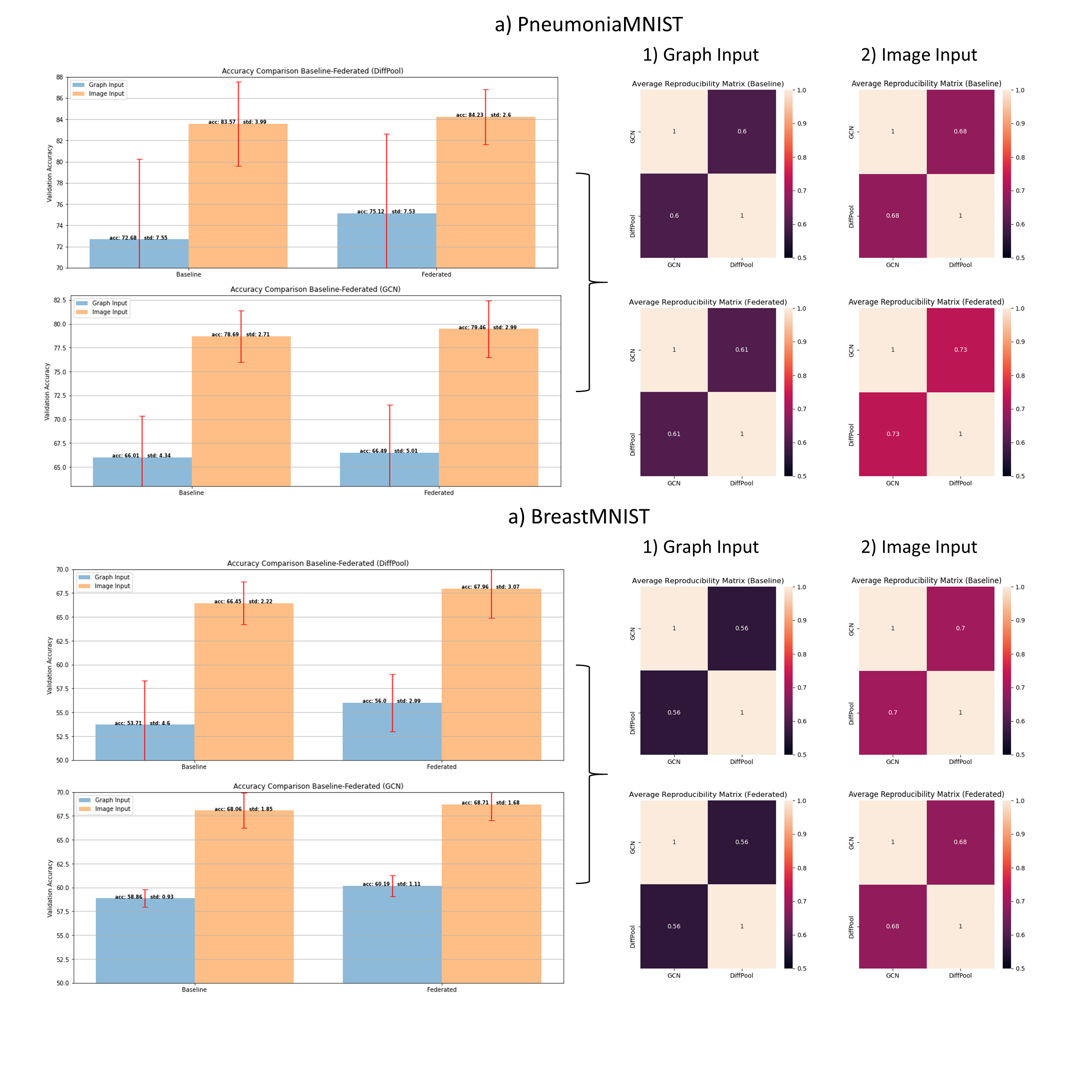}}
\caption{\emph{Accuracy and reproducibility score comparison of image-based and graph-based representations (biomedical image datasets only).} Bar charts show the accuracy comparison for both input representations using DiffPool and GCN models. 1) Graph Input $|$ the column shows the average reproducibility matrices of baseline and federated models when the input type is graph. 2) Image Input $|$ the column shows the average reproducibility matrices of baseline and federated models when the input type is image.}
\label{fig:supp}
\end{figure}

\section{Results and Discussion}

\textbf{Evaluation of biomedical image datasets.} We evaluated our federated reproducibility framework on two large-scale biomedical image datasets which are retrieved from MedMNIST\footnote{\url{https://medmnist.com/}} public dataset collection \cite{medmnistv2}. The first biomedical image dataset (PneumoniaMNIST dataset) contains 5856 X-ray images, with a size of $28 \times 28$, and belonging to a normal control class or displaying pneumonia which is a respiratory infection that affects the lungs \cite{gereige2013pneumonia}. Out of the 5856 subjects, we randomly selected 1000 samples with balanced classes (normal and pneumonia). The second dataset (BreastMNIST dataset) contains 780 breast ultrasound images, with the size of $28 \times 28$, belonging to a normal control or diagnosed with malignant breast cancer. We randomly sampled 546 subjects where 399 subjects are labeled as normal and 147 as malignant. We used two different representations of the imaging datasets to feed into the models. In the first representation, we simply fed the original image  to the target GNN whereas in the second representation we converted each image into weighted graph matrix. The weights of connectivity matrix were calculated using absolute differences in intensity between pixel pairs.

\textbf{Evaluation of connectomic datasets.} Additionally, we used the Autism Brain Imaging Data Exchange (ABIDE I) public dataset \cite{di2014autism} to evaluate our federated reproducibility framework on morphological brain networks \cite{soussia2018}. We used the left and right hemisphere brain connectivity datasets of autism spectrum disorder (ASD) and normal controls (NC). These datasets include 300 brain graphs with balanced classes. Both left and right hemispheres are parcellated into 35 regions of interest (ROIs) using Desikan-Killiany Atlas \cite{fischl2004sequence} and FreeSurfer \cite{fischl2012freesurfer} software. The connectivity weight encodes the average morphological dissimilarity in cortical thickness between two cortical ROIs as introduced in \cite{soussia2018,mahjoub2018brain}. 

\textbf{Pool of GNNs.} For our federated reproducibility framework, we used 2 state-of-the-art GNNs which are DiffPool \cite{ying2018hierarchical} and GCN \cite{kipf2016semi}. DiffPool includes a differentiable graph pooling module that is able to generate hierarchical representations of a given graph. Soft cluster assignments learned by DiffPool at each layer of GNN \cite{ying2018hierarchical} to capture the graph nested modularity. The original aim of GCN is to perform node classification. However, we adapted the original GCN to handle whole-graph-based classification as in \cite{nebli2022quantifying}. The code of \cite{nebli2022quantifying}\footnote{\url{https://github.com/basiralab/RG-Select}} was used to develop our framework.

\textbf{Training settings and hyperparameters.} To train models in a federated manner, we divided each dataset into $H = 3$ local (independent) sets. We also divided each local data into 3-folds where two folds are used for training and the left one for testing. We selected all of the learning rates empirically. For DiffPool, the learning rate is set to $10^{-4}$ across all datasets. For GCN, the selected learning rates are $10^{-6}$, $10^{-5}$, $10^{-5}$ and $5 \times 10^{-6}$ for the datasets PneumoniaMNIST, BreastMNIST, ASD/NC LH and ASD/NC RH, respectively. The threshold value $K$ for the top features is set to 20 in our experiment. The epoch size $E$ is fixed to 100 and batch size $B$ is set to 1. The number of communication rounds $C$ is set to 5.

\begin{figure}[ht!]
\captionsetup{belowskip=0pt}
\centerline{\includegraphics[width=13cm]{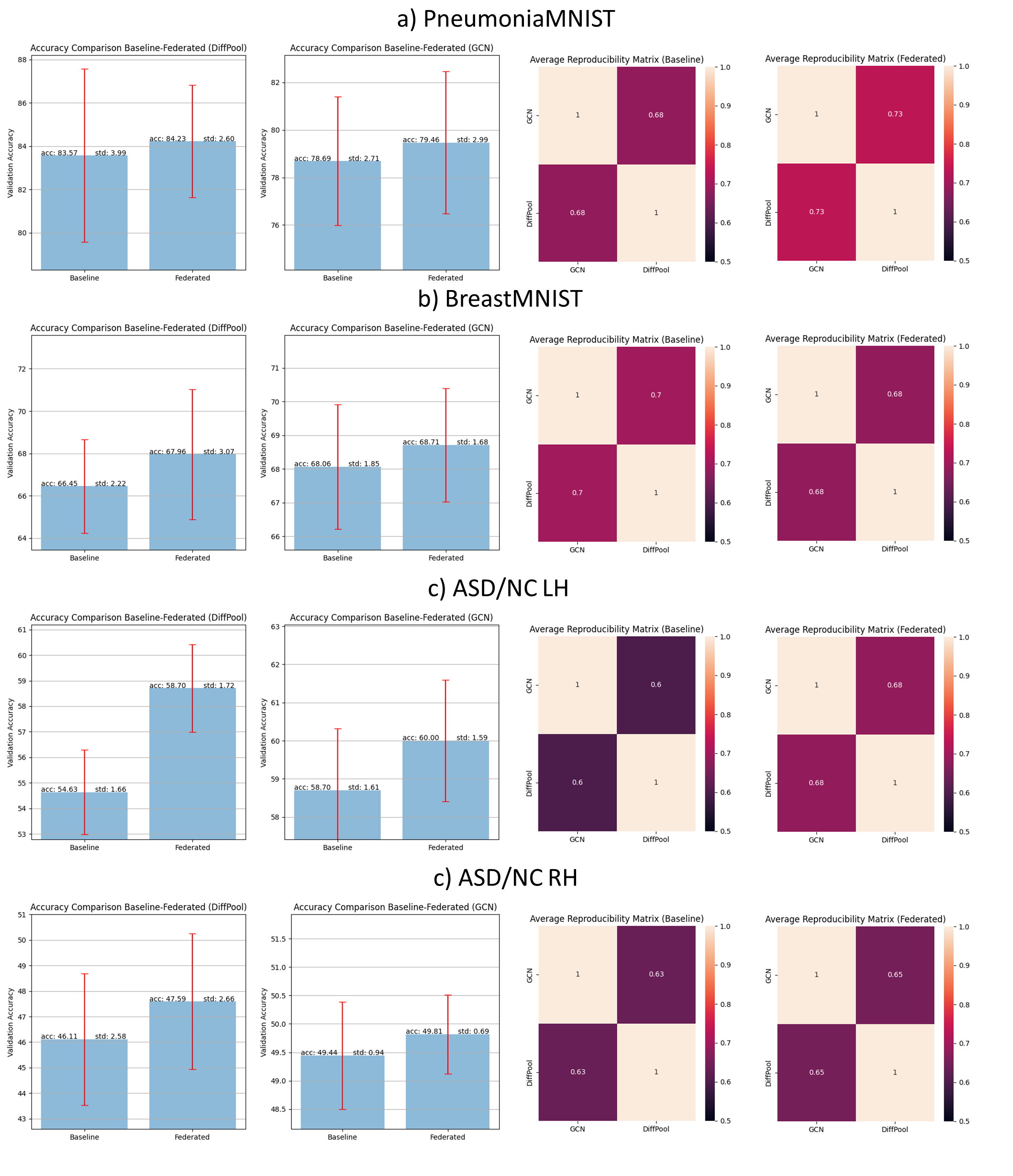}}
\caption{\emph{Accuracy and global reproducibility matrix comparison across datasets and GNN models.} Each row in the figure represents individual datasets. The first and second columns are the accuracy comparison results of DiffPool and GCN models, respectively. The third and fourth columns represent the baseline and federated reproducibility matrices, respectively. }
\label{fig:results}
\end{figure}

\textbf{Model accuracy and reproducibility evaluation.} We compared our federated reproducibility framework to the non-federated technique (without using \textbf{Algorithm}~\ref{FedAvgAlgorithm}). The comparison was performed for both validation accuracies and average reproducibility matrices storing the intersection ratio of the top $K$ discriminative biomarkers between global GNN models. \textbf{Fig}~\ref{fig:results} shows the comparison results of the classification accuracy and reproducibility matrices for two biomedical image datasets and two connectomic datasets. Notably, the classification accuracy was boosted across all datasets for each local model using federation. For the datasets, PneumoniaMNIST, ASD/NC LH and ASD/NC RH, an increase in the GNN reproducibility score is noted. However, a slight decrease was observed when we evaluated our federated reproducibility framework with the BreastMNIST dataset. The results of biomedical image datasets displayed in \textbf{Fig}~\ref{fig:results} were obtained when traininng GNN models on the original images directly. \textbf{Fig.}~\ref{fig:supp} displays the accuracy and reproducibility score comparison of the graph and image representations of the biomedical image datasets. Interestingly, according to \textbf{Fig.}~\ref{fig:supp}, models performed better in terms of both accuracy and reproducibility when the original images were used without resorting to transforming them into graphs. 

\begin{figure}[H]
\captionsetup{belowskip=0pt}
\centerline{\includegraphics[width=10cm]{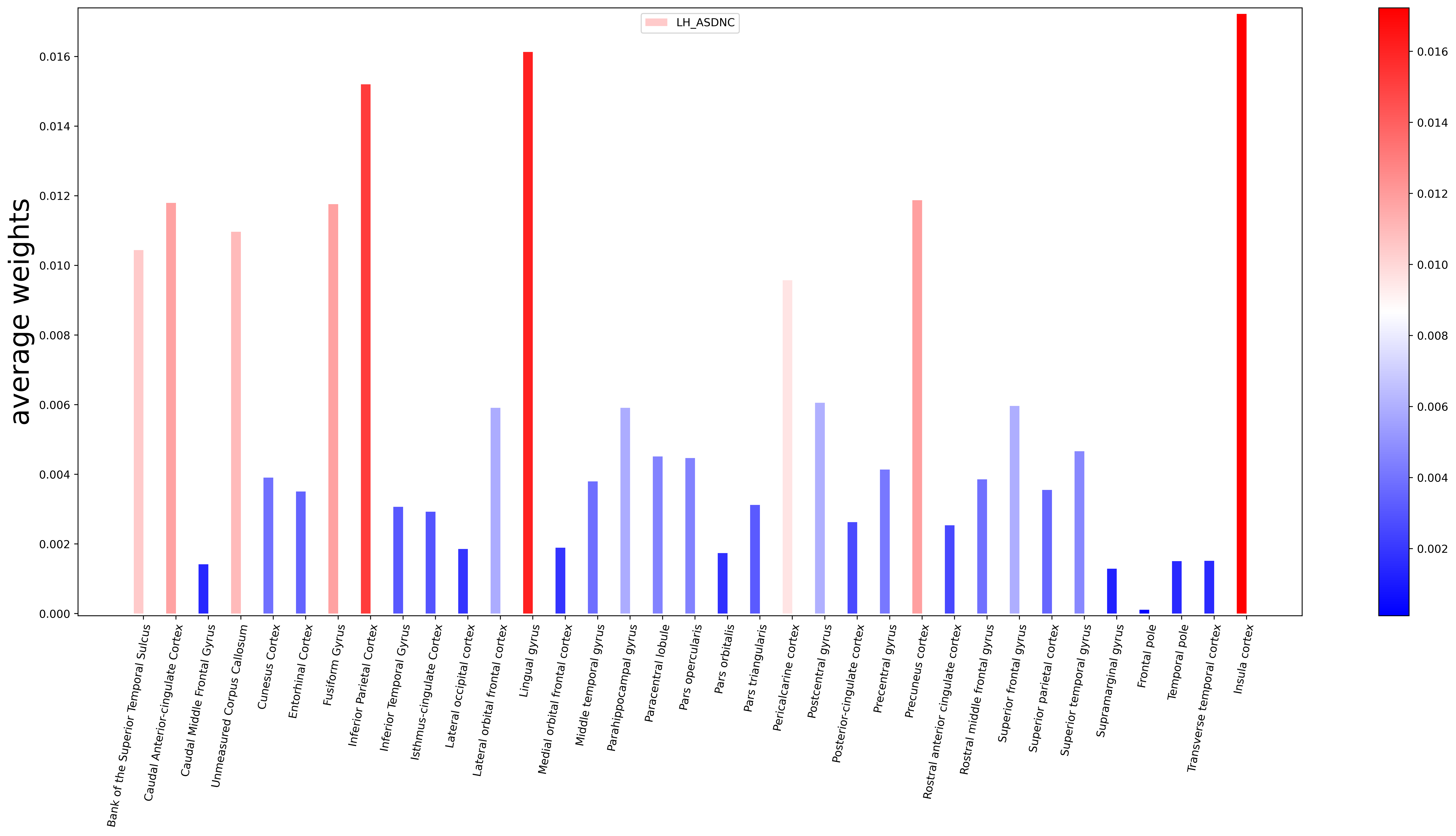}}
\caption{\emph{The learned weights of the cortical regions by the most reproducible GNN model for the dataset ASD/NC LH.} }
\label{fig:LH_regions}
\end{figure}

\begin{figure}[H]
\captionsetup{belowskip=0pt}
\centerline{\includegraphics[width=10cm]{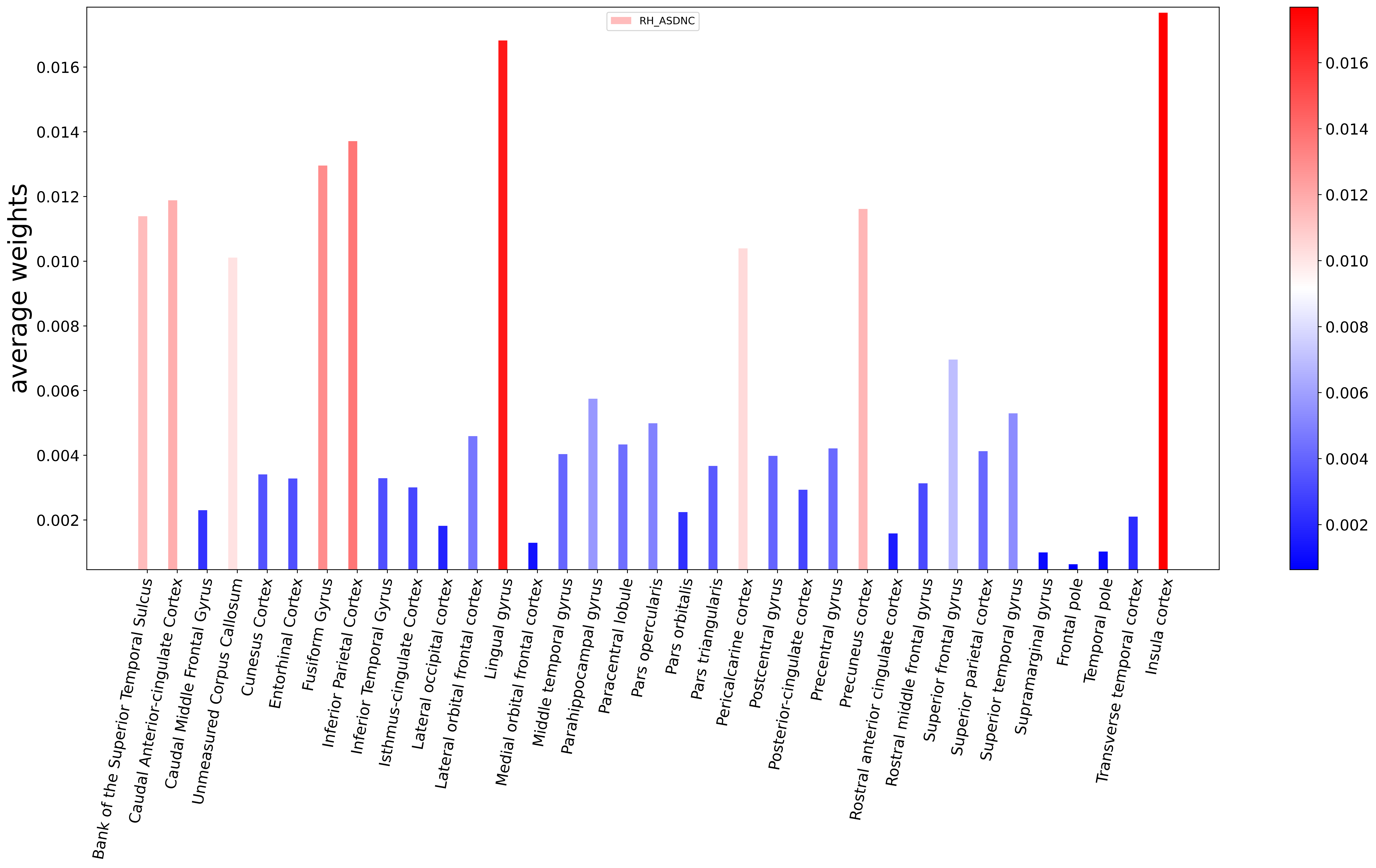}}
\caption{\emph{The learned GNN weights of the cortical regions by the most reproducible GNN model for the dataset ASD/NC RH.} }
\label{fig:RH_regions}
\end{figure}

\textbf{Most reproducible connectomic biomarkers.} \textbf{Fig}~\ref{fig:LH_regions} and \textbf{Fig}~\ref{fig:RH_regions} shows the absolute value of the feature weights learned by the globally most reproducible GNN, which are the averages of the locally learned weights using ASD/NC LH and RH datasets, respectively. We considered the global GNN model rather than the hospital-specific local models to select the most reproducible biomarkers since the most reproducible model may change across hospitals. According to \textbf{Fig}~\ref{fig:RH_regions}, the insula cortex and lingual gyrus are selected as the most reproducible biomarkers for both LH and RH datasets followed by the precuneus and the inferior parietal cortex. In patients presenting with autism, the insula cortex shows an important variation in T1 according to \cite{lou2021quantitative}. Such finding embodies the nature of this neurodevelopmental disorder mainly characterized by altered cognitive, emotional and sensory functions. These neurological aspects of the disease are orchestrated by the insular cortex \cite{gasquoine2014contributions} pinpointing further that autism is considered an insula pathology and highlighting the reliability of such biomarker as a fingerprint of the disease \cite{nomi2019insular}.
\cite{gebauer2015there} demonstrated a significant relationship between ASD traits and cortical thickness of the lingual gyrus. As a matter of fact, it has been linked to the specific aspect of sensory disturbances in ASD \cite{habata2021relationship}. Regarding the precuneus, the medial part of the posterior parietal lobe, it has been linked to a specific clinical phenotype of ASD which is associated with psychological comorbidities, such as post-traumatic stress disorder. According to \cite{kitamura2021association}, the reduction in the precuneus gray matter was correlated with adverse childhood experiences leading to intrusive reexperiencing in adults with ASD. Thus, the precuneus represents a potential biomarker of the disease even more valuable since it could be phenotype-dependant. Furthermore, the almost miror effect discernible by comparing both hemispheres (Fig 3 and 4) might be explained by the heterogeneity of the sample with patients' age ranging from 5 to 64 years (mean age of onset = 14 years). It pinpoints the evolving aspect of the morphological abnormalities over time going from being primarily left-lateralized to inter-hemispheric differences diminishing  progressively when reaching adulthood \cite{khundrakpam2017cortical}. 

\textbf{Limitations and future directions.} Even though we used different datasets to evaluate our federated reproducibility framework, it has several limitations. First,  we assumed that each local hospital has almost the same number of samples --which might not be the case for decentralized medical datasets.  Second, we only used 2 different GNNs. In our future work, we aim to optimize our hyperparameters using advanced methods, use an early stopping technique, consider imbalanced data distributions across hospitals and extend the pool of GNNs to obtain more results for an enhanced comparison and generalizability. Incorporating clinical features of patients such as a detailed  assessment of cognition, sensory disturbances and the presence of comorbidities may help  add phenotypic value to the already established biomarkers of the ASD in our study.

\section{Conclusion}
In this paper, we investigated and quantified the reproducibility of GNN models trained in a federated manner. We evaluated our federated reproducibility framework using several medical imaging and connectomic datasets. Our framework aims to calculate the most reproducible biomarkers or features while handling decentralized datasets and boosting the local model accuracies. In this prime work, we showed that federated learning not only increases the performance of locally trained GNN models but also boosts their reproducibility. In our future work, we will investigate federated GNN reproducibility when learning on non-IID clinical datasets and examine other state-of-the-art GNN models.

\section{Supplementary material}

We provide three supplementary items for reproducible and open science:

\begin{enumerate}
	\item A 7-mn YouTube video explaining how our framework works on BASIRA YouTube channel at \url{https://youtu.be/pnattjt981k}.
	\item Our code in Python on GitHub at \url{https://github.com/basiralab/reproducibleFedGNN}. 
	\item A GitHub video code demo on BASIRA YouTube channel at \url{https://youtu.be/bG54z0v75U0}. 
\end{enumerate}

\section{Acknowledgements}

This work was funded by generous grants from the European H2020 Marie Sklodowska-Curie action (grant no. 101003403, \url{http://basira-lab.com/normnets/}) to I.R. and the Scientific and Technological Research Council of Turkey to I.R. under the TUBITAK 2232 Fellowship for Outstanding Researchers (no. 118C288, \url{http://basira-lab.com/reprime/}). However, all scientific contributions made in this project are owned and approved solely by the authors.

\bibliography{Biblio3}
\bibliographystyle{splncs}
\end{document}